# Revealing Shadows: Low-Light Image Enhancement Using Self-Calibrated Illumination

Farzaneh Koohestani[1], Nader Karimi[1], Shadrokh Samavi[2,3]
[1]Department of Electrical and Computer Engineering Isfahan University of Technology, Isfahan, Iran
[2]Department of Computer Science Seattle University, Seattle, USA
[3]Department of Electrical and Computer Engineering, McMaster University, Hamilton, Canada

*Abstract*—In digital imaging, enhancing visual content in poorly lit environments is a significant challenge, as images often suffer from inadequate brightness, hidden details, and an overall reduction in quality. This issue is especially critical in applications like nighttime surveillance, astrophotography, and low-light videography, where clear and detailed visual information is crucial. Our research addresses this problem by enhancing the illumination aspect of dark images. We have advanced past techniques by using varied color spaces to extract the illumination component, enhance it, and then recombine it with the other components of the image. By employing the Self-Calibrated Illumination (SCI) method, a strategy initially developed for RGB images, we effectively intensify and clarify details that are typically lost in low-light conditions. This method of selective illumination enhancement leaves the color information intact, thus preserving the color integrity of the image. Crucially, our method eliminates the need for paired images, making it suitable for situations where they are unavailable. Implementing the modified SCI technique represents a substantial shift from traditional methods, providing a refined and potent solution for low-light image enhancement. Our approach sets the stage for more complex image processing techniques and extends the range of possible real-world applications where accurate color representation and improved visibility are essential.

## 1- Introduction

In computer vision, the deleterious effects of suboptimal lighting conditions on image quality cannot be overstated—particularly in critical applications spanning object detection, segmentation, surveillance, and medical imaging. Enhancing illumination in low-light images is thus a critical task, essential for revealing details shrouded in shadow. The task of revealing the obscured details within darkened images presents a formidable challenge, primarily due to the exceedingly low signal-to-noise ratio inherent in such conditions. Despite these difficulties, the wide-ranging implications for improvement in various computer vision applications have spurred intense focus within the research community. This concentrated effort has yielded significant breakthroughs in enhancing low-light images, reflecting the vigorous pursuit of advancements in the field [1].

In recent years, various methods have been introduced to address this classic problem, broadly categorized into two main approaches: traditional model-based methods and deep learning-based methods.

Traditional methods essentially treat the degradation occurring in an image as a model and attempt to enhance images by estimating the parameters of this model. However, these methods often possess limited capabilities and rely heavily on expert understanding to consider many factors in constructing the initial model [2]. They cannot learn these aspects directly from the data. The Retinex theory, as outlined in foundational literature [4], provides a principal framework for low-light image enhancement by positing that an image under low illumination can be decomposed into two components: illumination and reflectance (the latter equating to a clearer image).

Conversely, methods within the second category — those based on deep learning — adopt a distinct perspective. They leverage the power of deep neural networks to extract the necessary information from extensive training datasets. This approach enables the enhancement process to be conducted with a more informed and adaptive methodology. Deep learning algorithms are designed to discern intricate patterns autonomously and features instrumental for low-light image enhancement, thus often outperforming their traditional counterparts by learning directly from the data [3], [4].

Low-light image enhancement has seen the advent of numerous sophisticated techniques to overcome the challenges associated with inadequate lighting and noise, all while retaining critical detail and true-to-life colors. The LIME method distinguishes itself by estimating the illumination for each pixel [3], thereby shedding light on the darker regions of an image. RetinexNet adopts a kindred philosophy, utilizing deep learning to distinguish and independently adjust the illumination and reflectance within an image [4]. DeepUPE harnesses convolutional neural networks to learn from examples of well-exposed images, automatically calibrating the exposure of underlit photos for a well-balanced outcome. Addressing the issue of resolution degradation, Zero-DCE presents a curve estimation approach to brighten low-light images efficiently without relying on deep learning models [6].

On the other hand, RRDNet stands out by restoring resolution and reducing noise [7], particularly in challenging lighting scenarios. At the same time, the STAR technique impressively preserves structural and textural integrity during illumination adjustment [8].

EnlightenGAN [9] leverages generative adversarial networks to enhance poorly lit images without paired training data, and the Self-Calibrated Illumination method advances this with selective illumination enhancement [10], preserving color integrity. SID [11] pushes the boundaries by processing raw images to boost visibility in extremely low-light conditions and smooth out noise without losing detail. DeepLPF applies learned filtering [12]. Lastly, the RF method employs reinforcement learning to control image editing software autonomously [13], optimizing unpaired image enhancement in a novel way. Each method contributes to the evolving landscape of image processing, offering solutions tailored for various scenarios, from consumer photography to professional applications where clarity in low light is paramount.

Segmenting low-light images into distinct components is a strategic approach to ameliorate poor lighting conditions. One can meticulously calibrate the exposure in underlit and overlit regions by isolating and enhancing the luminance component. Merging this refined luminance back with the original color information and intricate details allows for the reassembly of the image. This process results in a marked improvement in the image's overall quality and visual clarity, specifically tailored for low-light environments.

Unsupervised learning is propelling progress across numerous disciplines, notably in contexts where precise ground truth is scarce. This method enables training models and algorithms without requiring direct reference data annotation. In environments where obtaining ground truth proves difficult or expensive, unsupervised learning capitalizes on the inherent structures and patterns present within the data. Such a method fosters advancements in a wide array of fields, even when faced with limited access to detailed and accurate ground truth.

This paper presents a method focused on enhancing image brightness by decoupling the brightness component from the color components. By isolating and exclusively refining the brightness aspect through a rapid and robust unsupervised technique, we aim to uplift the image's luminance significantly. The enhanced brightness component is then reinserted into the original dark image, replacing the older, dimmer brightness levels. This approach ensures that the improved image exhibits increased brightness and retains the integrity of the original color information, resulting in a well-lit and color-accurate visual output.

## 2- Proposed Method

Photography and image processing challenges are greatly amplified under low-light conditions, which can critically degrade image quality. When capturing images in environments with inadequate lighting, cameras often struggle to record the finer details, leading to increased noise, motion blur, and poor color fidelity. These problems are exacerbated in digital photography, where the camera's sensor may not be sensitive enough to distinguish between subtle shades of light, rendering murky images that lack clarity. This loss of information, particularly in the illumination details, compromises the image's overall integrity.

### A. *Illumination Component Extraction*

The degradation of image quality in low-light conditions is often attributed to the loss of illumination information. A promising approach to mitigate this is to extract and enhance the brightness component of an image. By amplifying the luminance data, one can potentially recover details that are obscured in the shadows, thereby improving the visibility and clarity of the image. However, this process is complicated because most digital images are captured and stored in the RGB (Red, Green, Blue) color space, where the brightness information is inherently intertwined with color information. In the RGB model, each color channel contributes to the image's overall brightness, making it challenging to isolate and adjust the luminance without affecting the color balance [3].

Color space transformations, while a long-standing concept in image processing, have not been widely utilized to enhance low-light images. Our approach harnesses the distinct separation of luminance and chrominance within color spaces such as YUV, Lab, or HSV. By tapping into this characteristic, we've developed a technique that improves images captured in poorly lit environments by precisely adjusting the light information, thus maintaining the integrity of the image's color data.

In this method, we have applied color space transformations in a novel way to address the persistent issue of inadequate lighting in photography and videography, providing a solution that selectively enhances the luminance component without altering the chromatic components. This targeted enhancement is key to achieving well-lit and color-accurate images, reflecting the original scene with greater fidelity and clarity.

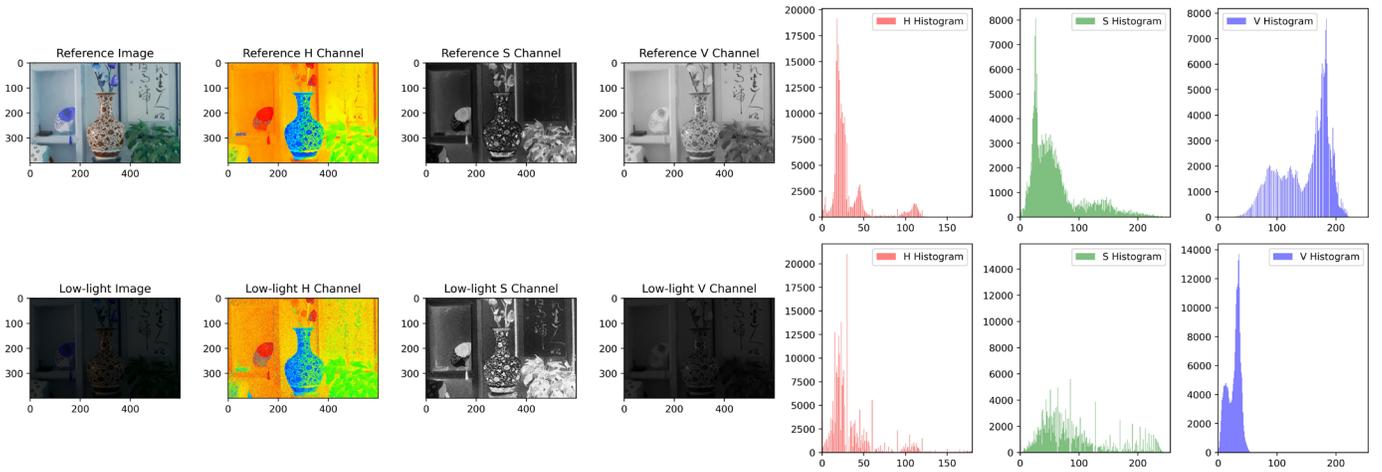

Fig. 1: Analysis of the brightness and color components within the HSV color space for a selected image from the LOL dataset [14]. The top row displays a reference image with distinct H (Hue), S (Saturation), and V (Value) channels, accompanied by histograms demonstrating an even distribution across all components. In contrast, the bottom row presents the underexposed image revealing a compressed V channel, indicative of diminished brightness, while the H and S channels remain relatively unaffected.

When an image is captured in a standard RGB color space, the light and color information are intermingled, which can complicate efforts to enhance one aspect without inadvertently affecting the other. Transforming the image into a color space like HSV allows for individual manipulation of the 'V' (value) channel that represents brightness, enabling precise adjustments to the image's luminance. The Value (V) in the HSV color space, representing the brightness of the color, is calculated from the normalized RGB components as follows:

$$V = \max(R, G, B) \quad (1)$$

where R, G, and B represent the color's red, green, and blue components.

Likewise, the YCbCr color space separates the luma (brightness) from the chroma (color) components. The 'Y' channel is dedicated to luma, representing the brightness level distinct from the 'Cb' and 'Cr' chroma channels. The calculation of the Y component (luminance) in the YCbCr color space from the RGB components is formulated as:

$$Y = 0.299R + 0.587G + 0.114B \quad (2)$$

where $R$, $G$, and $B$ are the red, green, and blue components of a pixel's color in the RGB color space, respectively. This separation is essential for image processing tasks that require adjustments to luminance without affecting color balance.

Figures 1 and 2 from the LOL dataset [14] showcase the significance of extracting and enhancing the brightness component in images to improve underexposed (dark) areas. Figure 1 analyzes an image within the HSV space, where the top row's well-lit image has clear H, S, and V histograms, while the bottom row's underexposed image reveals a squeezed V (brightness) histogram, highlighting the need for brightness enhancement. Figure 2 observes the YCbCr space, contrasting a reference image with even Y, Cr, and Cb histograms against a low-light image with a constricted Y (luminance) histogram. This comparison illustrates the necessity of luminance improvement to recover detail and enhance image quality in low-light conditions.

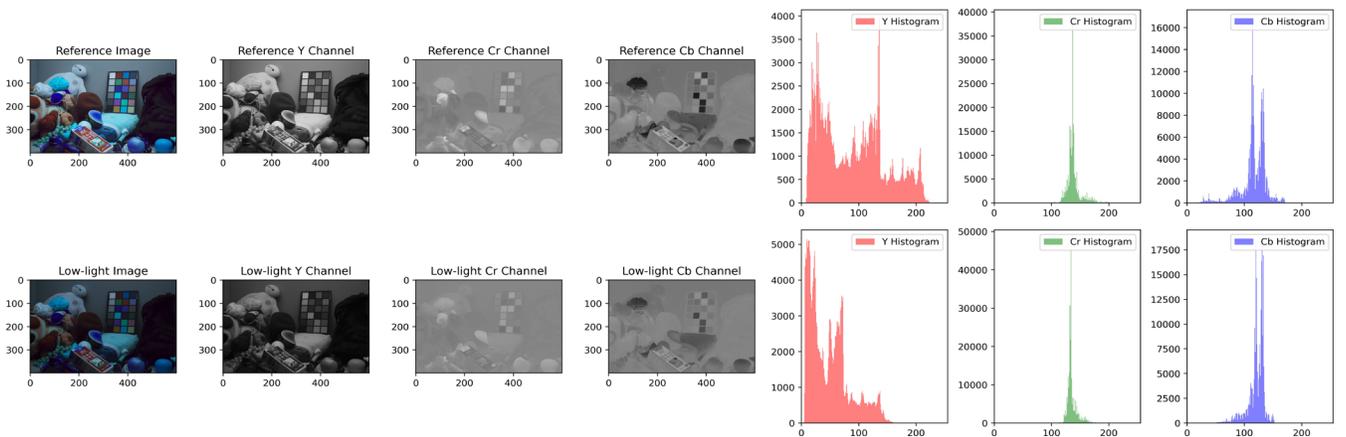

Fig. 2: Investigation of the luminance and color channels in the YCbCr color space within a selected image from the LOL dataset [14]. Top row: Reference image with detailed Y, Cr, Cb channels and balanced histograms. Bottom row: Low-light image with a compressed Y channel, approximately consistent Cr, Cb channels, and histograms illustrating the need for luminance enhancement.

The luminance channel becomes the focal point for enhancement. Techniques such as curve adjustments can remap the brightness levels, making dark regions more discernible. Contrast stretching is another technique that can widen the range of tonal values, revealing details from the shadows. Adaptive histogram equalization is particularly useful for its ability to adjust the contrast on a local scale, which can benefit images with uneven lighting.

The chrominance channels are left untouched throughout this process, which is key to maintaining the image's original hues and saturation levels. This focused approach to enhancement avoids the unintended color shifts that can result from more indiscriminate processing methods.

After the luminance channel has been enhanced, the image is transformed back into the RGB color space, which effectively merges the improved brightness with the original color data. The resulting image is not only brighter and clearer in areas that were previously underexposed but also retains the color fidelity of the original scene. This characteristic makes color space transformation a valuable tool in various contexts, including digital photography, where it is essential to capture the mood of a scene realistically, and video production, where consistent color representation is crucial across different lighting conditions. This method ensures that crucial details are visible in security and surveillance, even in footage captured in low-light environments.

While color space transformations are well-established in image processing, their application in low-light image enhancement continues to be a reliable approach to improving image visibility without sacrificing the authenticity of the original colors. This method offers a disciplined balance between enhancing light information and preserving color, ensuring images remain true to the viewer's perception of the natural scene.

### B. *Self-Calibrated Module*

Unsupervised methods for enhancing low-light images offer significant advantages, particularly their independence from paired sample data. Unsupervised techniques enable a more flexible and scalable approach to image enhancement by eliminating the need for matched low and highlight images. This advantage is invaluable in medical imaging and surveillance, where acquiring perfectly paired images under different lighting conditions is often impractical or impossible. In medical applications, for example, enhancing images without a reference standard allows for clearer visualization of diagnostic imagery, which is crucial for accurate analysis and treatment planning. Similarly, unsupervised methods provide a robust solution to improve image clarity and detail in monitoring scenarios where lighting conditions can be unpredictable and controlled samples are unavailable. The use of unsupervised learning, therefore, represents a critical step forward in processing images from environments where obtaining a modified sample of the low-light conditions would be challenging or unfeasible.

In this paper, we employ the Self-Calibrated Il-lumination learning framework [10]—a state-of-the-art method characterized by its speed, flexibility, and robustness in enhancing images under challenging low-light conditions. The SCI framework operates on a cascaded illumination learning process. The process incorporates weight sharing, a design choice that streamlines the enhancement process and reduces computational complexity. To address the issue of color consistency with the SCI method, we've crafted a new technique that stays more faithful to the original image's colors. The main steps of this technique are outlined in the flowchart in Figure 3.

Central to the SCI method is the innovative use of a self-calibrated module. This module ensures the convergence of the enhancement results at each stage, allowing the framework to operate effectively with just one primary block during inference. This approach significantly reduces computational expenses compared to previous methods that relied on multiple vectors, presenting a more efficient solution.

Without modifying the original SCI model, this article adopts it in its entirety, leveraging its intrinsic capacity to adapt to different scenes unsupervised. The unsupervised training loss, a cornerstone of the SCI framework, bolsters the model's adaptability, enabling it to generalize to a wide range of real-world scenarios.

We apply the SCI method as is but conduct thorough investigations to unearth the inherent properties of the SCI framework that set it apart from prior work. These properties include an operation-insensitive adaptation, which guarantees stable performance across various operational settings, and a model-irrelevant generality, which allows for applying the SCI framework to existing models for brightness enhancement, thereby improving their efficacy. By utilizing the SCI method without alteration, this article showcases its robust capabilities and potential to enhance low-light images in practical applications significantly.

The Self-Calibrated Illumination learning framework, as detailed in [10], represents a significant advancement in the domain of low-light image enhancement due to its unique self-calibrated nature. This innovative approach employs a cascaded illumination learning process that shares weights across different stages, leading to a harmonious convergence of results at each stage. The self-calibrated module is the cornerstone of this methodology, enabling the system to stabilize exposure effectively while simultaneously minimizing computational overhead. By leveraging this self-calibration, the SCI framework can operate with a singular basic block during inference, markedly reducing the computational cost without sacrificing performance. This self-calibrated design enhances the framework's efficiency and maintains consistent visual quality, making it particularly well-suited for real-world applications where speed and flexibility are of the essence.

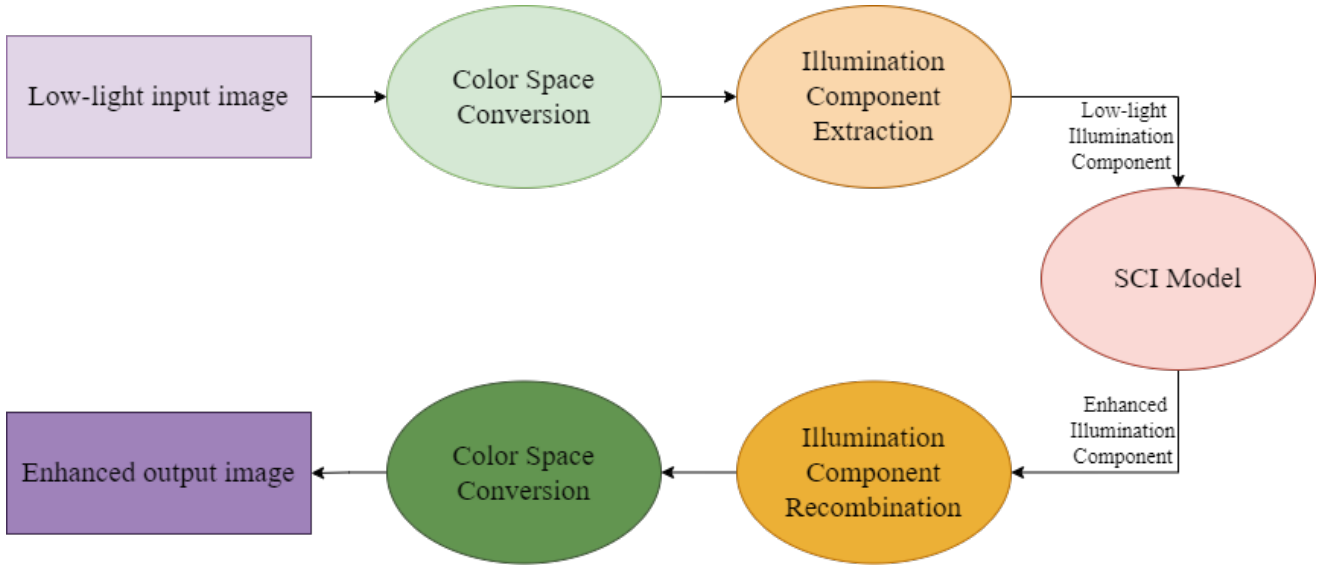

Fig. 3: Flowchart of our method, illustrating the low-light image enhancement process using SCI: starting from the import of the original RGB image, conversion to another color space that can extract the illumination component, application of the SCI enhancement techniques to the illumination component, and recombination with the other original components to produce the final enhanced image.

The Self-Calibrated Illumination learning framework represents a novel approach to enhancing low-light images through an unsupervised training process. This framework circumvents the need for paired datasets—typically a necessity in supervised learning methods—by using an internal mechanism to improve image quality iteratively. The unsupervised training loss is a composite of several components designed to address specific challenges associated with low-light enhancement. It ensures that images retain natural colors, structural details, and consistency throughout the enhancement stages while also effectively managing noise that tends to be amplified when brightening dark images. Moreover, the self-calibration aspect of the framework allows the model to refine its parameters recursively, leading to continuous improvements in performance.

The SCI method shows promise for enhancing low-light images, but it tends to process the entire RGB image without differentiating between brightness and color, which isn't always ideal. By enhancing only the lighting component of an image and keeping the color components intact, we've found a more effective way to improve image brightness.

Our refined approach targets this specific problem by isolating the lighting component. By focusing enhancement efforts on this aspect, we can significantly improve the visibility and sharpness of the image without distorting the original color palette. This is a crucial step because color components often don't need the same correction level as the lighting component in low-light images.

After the lighting component has been enhanced using our SCI-based technique, it is recombined with the untouched color components. This recombination is a delicate process that ensures the improved brightness blends seamlessly with the original colors, preserving the natural look of the image. The advantage of this targeted enhancement is twofold. First, it ensures that the image's colors remain true to life, especially in contexts where color fidelity is critical, such as in medical imaging or forensic photography. Second, it improves the overall perception of the image by selectively addressing the areas most affected by poor lighting conditions.

In essence, our method capitalizes on the strengths of the SCI technique while introducing a more nuanced application that respects the integrity of the original image's colors and focuses on the key issue of inadequate lighting. This leads to enhanced images that are brighter and clearer and maintain the authenticity of their original appearance.

## 3- Experimental Results

### A. Image datasets

Our experimental evaluation utilized two widely recognized paired datasets in the field: the LOL dataset [14] and the LOL- v2 Real dataset [15]. The LOL dataset comprises 500 pairs of low/highlight images, of which 485 pairs were designated for training and 15 pairs for testing. Additionally, 85 images were randomly selected from the training set to serve as a validation set. The LOL real dataset contains a larger pool of images, with 689 allocated for training and 100 for testing. One hundred eighty-eight images were randomly chosen from this dataset to form the validation set.

Despite these datasets featuring pairs of images with different lighting conditions, we focused solely on the low-light images for training our model. The corresponding well-lit reference images were not used in the training phase; instead, they served as a benchmark to assess the performance of our model post-enhancement. This evaluation method allowed us to measure how effectively our model could reconstruct the brightness and clarity of the underexposed images to match the quality of their well-lit counterparts, ensuring that our model's enhancements are practical and grounded in real-world comparison.

### B. Implementation Details

In developing our image enhancement model, we conducted our computations on an Nvidia RTX 3060, ensuring efficient handling of our deep learning tasks. All images used in training and validation were standardized to a resolution of 400 x 600 pixels. The model's parameters were optimized using the ADAM optimizer, chosen for its robustness, with a fine-tuned learning rate of 0.0003. We set the batch size to one to ensure detailed learning from our data. Although the training was assigned to a maximum of 1000 epochs, we incorporated early stopping, guided by the validation loss to halt training preemptively if no improvement was detected, effectively avoiding overfitting.

### C. Comparison with State-of-the-art

To demonstrate the effectiveness and generalizability of our method, we conducted a comparative analysis using two datasets—LOL [14] and LOL-v2 Real [15]—against several state-of-the-art methods. We employed two widely accepted quantitative criteria for this comparison: Peak Signal-to-Noise Ratio (PSNR) and Structural Similarity Index Measure (SSIM).

By employing HSV (Hue, Saturation, Value) and YCbCr (Luma, Blue-difference Chroma, Red-difference Chroma) color spaces, we could isolate and enhance only the illumination component of the images, leaving the color information intact. This separation is significant as it allows for targeted image brightness enhancement without altering the color balance, which is especially important in low-light image enhancement.

The results of our method, when utilizing these color spaces, were laid out in Table 1 for the LOL [14] dataset.

Table 1: Quantitative results (PSNR, SSIM) of state-of-the-art methods and ours on the LOL dataset [14]. "T", "S", and "U" represent "Traditional", "Supervised", and "Unsupervised" methods, respectively. The best result is in **red**, and the sub-optimal result is in **blue**.

| Method | Type | PSNR (dB) | SSIM |
|---|---|---|---|
| STAR [8] | T | 12.91 | 0.518 |
| RetinexNet [4] | S | 13.1 | 0.429 |
| DeepUPE [5] | S | 13.04 | 0.483 |
| LIME [3] | U | 14.92 | 0.516 |
| Zero-DCE [6] | U | **15.51** | **0.553** |
| RRDNet [7] | U | 11.4 | 0.457 |
| EnlightenGan [9] | U | **15.64** | **0.578** |
| SCI [10] | U | 14.78 | 0.522 |
| **Ours (HSV)** | U | 15.22 | 0.457 |
| **Ours (YCbCr)** | U | **16.06** | **0.542** |

Here, we compared our approach with other advanced methods, including LIME [3], RetinexNet [4], DeepUPE [5], Zero-DCE, RRDNet [7], STAR [8], EnlightenGan [9], and SCI [10]. Our method demonstrated superior performance in the LOL dataset benchmark, achieving the highest PSNR among the considered techniques and ranked as the third best in terms of the SSIM criterion. In Table 2, with the LOL-v2 Real dataset as a benchmark, our method proved to be the most effective, achieving the highest scores in both PSNR and SSIM compared to the SID [11], DeepUPE [5], DeepLPF [12], and RF [13] methods. This comparative analysis further reinforced the efficacy of our method in enhancing low-light images across different datasets and against various methods.

Table 2: Quantitative results (PSNR, SSIM) of state-of-the-art methods and ours on the LOL–v2 Real dataset [15]. "S", "R" and "U" represent "Supervised", "Reinforcement Learning," and "Unsupervised" methods, respectively. The best result is in **red**, and the sub-optimal result is in **blue**.

| Method | Type | PSNR (dB) | SSIM |
|---|---|---|---|
| SID [11] | S | 13.24 | 0.442 |
| DeepUPE [5] | S | 13.27 | 0.452 |
| DeepLPF [12] | S | **14.10** | 0.48 |
| RF [13] | R | 14.05 | **0.458** |
| **Ours (HSV)** | U | **16.64** | **0.518** |
| **Ours (YCbCr)** | U | **17.09** | **0.567** |

Overall, using PSNR as evaluation criteria provided a comprehensive understanding of our method's performance in error minimization, affirming its potential as a robust solution for low-light image enhancement. Following rigorous testing against various advanced methods using the LOL and LOL-v2 Real datasets, our method has emerged as the new state-of-the-art in low-light image enhancement.

## 4- Conclusion

In conclusion, our research aimed to enhance low-light images by adapting the Self-Calibrated Illumination (SCI) method to a color space that allows for the isolation of the luminance component. This strategy effectively reduced noise and color distortion, common challenges in enhancing low-light images. We implemented an early stopping mechanism to combat the overfitting problem often seen in the SCI model, contributing to better generalization and performance. Our experiments, conducted using the LOL and LOL-v2 Real datasets, showcased the superiority of our method, with our results surpassing those of several established techniques in the field, as evidenced in Table 2. Moreover, this work presents a robust solution for low-light image enhancement and lays the groundwork for future research to improve further and build upon these methodologies. Through continued exploration and refinement, we anticipate the development of even more sophisticated techniques that can handle a more comprehensive array of low-light conditions with greater effectiveness. The proposed method can be used as a preprocess for medical images [16], [17] and saliency detection [18], [19] algorithms.